\documentclass[lettersize,journal]{IEEEtran}
\usepackage{amsmath,amsfonts}
\usepackage{algorithmic}
\usepackage{algorithm}
\usepackage{array}
\usepackage[caption=false,font=normalsize,labelfont=sf,textfont=sf]{subfig}
\usepackage{textcomp}
\usepackage{stfloats}

\usepackage{placeins}
\FloatBarrier

\usepackage{url}
\usepackage{verbatim}
\usepackage{graphicx}
\usepackage{cite}
\usepackage{amssymb}
\usepackage{color}
\usepackage{rotating}
\usepackage{tabularx,caption}
\usepackage{booktabs,multirow,array}
\usepackage{siunitx} % For aligning numbers by decimal point
\usepackage{xurl}
\begin{document}

\title{Imbalance-Aware Culvert-Sewer Defect Segmentation Using an Enhanced Feature \\Pyramid Network}

\author{IEEE Publication Technology,~\IEEEmembership{Staff,~IEEE,}

\author{
    \IEEEauthorblockN{Rasha Alshawi\IEEEauthorrefmark{1}, Md Meftahul Ferdaus\IEEEauthorrefmark{1}, Mahdi Abdelguerfi\IEEEauthorrefmark{1}, Kendall Niles\IEEEauthorrefmark{2}, Ken Pathak\IEEEauthorrefmark{2}, Steve Sloan\IEEEauthorrefmark{2} }
    \IEEEauthorblockA{\IEEEauthorrefmark{1}University of New Orleans, New Orleans, Louisiana, USA}
    \IEEEauthorblockA{\IEEEauthorrefmark{2}US Army Corps of Engineers, Vicksburg, Mississippi, USA}
    \IEEEauthorblockA{Emails: \{rralshaw, mferdaus, gulfsceiDirector\}@uno.edu, \{Kendall.N.Niles, ken.pathak, steven.d.sloan\}@erdc.dren.mil}
}

%\author{Rasha Alshawi, Md Meftahul Ferdaus, Kendall Niles, Mahdi Abdelguerfi
        % <-this % stops a space
    
\thanks{This paper was produced by the IEEE Publication Technology Group. They are in Piscataway, NJ.}% <-this % stops a space
\thanks{}}

\markboth{Journal of \LaTeX\ Class Files,~Vol.~14, No.~8, August~2024}%
{Shell \MakeLowercase{\textit{et al.}}: A Sample Article Using IEEEtran.cls for IEEE Journals}


\maketitle

\begin{abstract}
Imbalanced datasets are a significant challenge in real-world scenarios. They lead to models that underperform on underrepresented classes, which is a critical issue in infrastructure inspection. This paper introduces the Enhanced Feature Pyramid Network (E-FPN), a deep learning model for the semantic segmentation of culverts and sewer pipes within imbalanced datasets. The E-FPN incorporates architectural innovations like sparsely connected blocks and depth-wise separable convolutions to improve feature extraction and handle object variations. To address dataset imbalance, the model employs strategies like class decomposition and data augmentation. Experimental results on the culvert-sewer defects dataset and a benchmark aerial semantic segmentation drone dataset show that the E-FPN outperforms state-of-the-art methods, achieving an average Intersection over Union (IoU) improvement of 13.8\% and 27.2\%, respectively. Additionally, class decomposition and data augmentation together boost the model's performance by approximately 6.9\% IoU. The proposed E-FPN presents a promising solution for enhancing object segmentation in challenging, multi-class real-world datasets, with potential applications extending beyond culvert-sewer defect detection.
\end{abstract}


\section{Introduction}

\IEEEPARstart{C}{omputer} vision has transformed industries by allowing machines to analyze visual data. A key aspect of this technology is semantic segmentation, which classifies individual image pixels into predefined categories \cite{hyun2021adjacent,peng2024hsnet}. This capability is vital for infrastructure maintenance, particularly in identifying structural elements like culverts and sewer pipes. Culverts and sewer pipes are essential water management structures \cite{iqbal2023scaled,wang2024automatic} that need regular inspection for damage like cracks, holes, and encrustation. Traditional inspection methods, like video pipe inspection, rely on manual review of footage, which is time-consuming and susceptible to human error \cite{adams2024national}. Automated semantic segmentation techniques can enhance inspection accuracy and efficiency. Advanced computer vision algorithms enable timely deficiency detection and repair, improving infrastructure integrity and longevity.

Segmenting culverts and sewer pipes is challenging due to their varying shapes, sizes, and environmental conditions. Despite advancements in semantic segmentation, accurately identifying these structures is difficult \cite{kuchi2020levee}. The small size and subtle appearance of many defects complicate the task, requiring models to capture fine details while understanding the overall pipe structure \cite{Li2023, Oh2022}. Factors such as diverse appearances, occlusions from vegetation or debris, and inconsistent lighting conditions can significantly impact the performance of current segmentation models \cite{Yi2022, Wang2021, Xia2022, He2022, Rius2022}.

Moreover, the scarcity of publicly available datasets for culvert and sewer pipe inspection complicates the development and testing of effective models. This limits the ability to train models that can generalize well to the wide variety of conditions encountered in real-world conditions.

Our study involved collecting a specialized dataset for culvert and sewer pipe inspection in response to this gap. The dataset revealed a significant imbalance in defect types, with common defects like cracks or joint misalignments being overrepresented, while rarer but critical defects, such as holes or collapses, are underrepresented. This imbalance can lead to models that perform well on frequent issues but struggle with less common, yet critical, structural problems \cite{li2024deep}.

Addressing these challenges is crucial for developing reliable automated systems for infrastructure inspection and maintenance. Effective segmentation enables precise detection of deficiencies, ensuring timely repairs.

Deep learning models like U-Net \cite{fan2023saca}, Feature Pyramid Network (FPN) \cite{panboonyuen2021transformer}, and Vision Transformers (ViT) \cite{bosco2023deep, chalcroft2023large} have been used for semantic segmentation tasks.
 These models have shown promising results in general segmentation tasks but struggle with challenges posed by culverts and sewer pipes. For example, U-Net and FPN may lack robustness to handle appearance variability, while ViTs, although powerful, can be computationally expensive and require extensive training data.

Recent advances in deep learning show promise in overcoming these challenges, particularly in improving FPNs \cite{Wu2021, Quyen2023, seferbekov2018feature, lin2017feature,lu2023new}. FPNs use a hierarchical pyramid of feature maps to capture information at multiple scales, improving the detection of large pipe features and small defects in culvert and sewer systems. This multi-scale approach is crucial for infrastructure inspection, where various feature scales must be accurately identified. However, while FPNs address the multi-scale issue, they do not inherently resolve the class imbalance in real-world inspection data.

Comprehensive solutions for managing object variations and addressing data imbalance in defect detection are needed. This involves adapting existing architectures for underground infrastructure inspections, developing new loss functions for class imbalance, and implementing advanced data augmentation techniques to represent underrepresented defect classes. These innovations are crucial for creating robust, reliable automated inspection systems for culverts and sewers, leading to more efficient maintenance and improved infrastructure health assessment \cite{Toan2023, Hu2021, Ji2023}.

In this paper, we introduce an enhanced FPN (E-FPN), a novel architecture for semantic segmentation in imbalanced culvert and sewer datasets. The E-FPN builds on the traditional FPN by incorporating enhancements to better handle object variations and improve feature extraction, addressing both the multi-scale challenges and the issue of class imbalance in underground infrastructure inspection. Our work makes two key contributions:
\begin{itemize}
    \item A customized E-FPN architecture for semantic segmentation of culverts and sewer pipes in imbalanced datasets:
    \begin{enumerate}
        \item We introduce a sparsely connected block for efficient information flow.
        \item We use depth-wise separable convolutions to reduce parameters without sacrificing representational power.

    \end{enumerate}
        These architectural innovations reduce computational complexity while maintaining, and often improving, segmentation performance.
    \item Exploring and validating techniques to mitigate data imbalance and enhance model performance:
    \begin{enumerate}
       \item We implemented class decomposition by partitioning the dataset into smaller, more homogeneous groups based on defect characteristics and sample distribution. This strategy allowed the model to focus on learning features specific to each type of defect more effectively. After training individual models on these smaller groups, we combined their predictions using ensemble learning techniques, enabling the final model to leverage the strengths of each sub-model. This approach enhanced the overall prediction accuracy, particularly for underrepresented classes.

        \item We use data augmentation to expand and balance the dataset, increasing diversity and ensuring fair representation of defect classes during training. 
    \end{enumerate}
    
\end{itemize}

Our analysis demonstrates the effectiveness of our methods for detecting defects in culverts and sewer systems. We tested our model on a diverse real-world dataset with nine classes of pipe images, varying in material, size, and orientation. To assess its versatility, we evaluated the model on a benchmark aerial semantic segmentation drone dataset, presenting unique challenges due to varying altitudes, perspectives, and environmental conditions. The E-FPN model demonstrated exceptional performance and adaptability across both datasets, showcasing its potential for addressing real-world semantic segmentation challenges beyond underground infrastructure inspection.

The paper is structured as follows: Section \ref{sec:RWork} reviews the evolution of semantic segmentation techniques, focusing on the development of FPN and its applications. Section \ref{sec:method} details our methodology, including the design and implementation of our proposed E-FPN architecture. Section \ref{sec:experiments} presents the Culvert-Sewer Defects dataset and the benchmark Aerial Semantic Segmentation Drone dataset used to evaluate the model's performance. Section V discusses techniques for balancing the dataset. Section VI outlines the implementation steps and metrics. Section VII provides an analysis of our experimental results and comparative evaluations against state-of-the-art methods. Finally, Section VIII concludes the paper with a summary of our contributions and potential future research directions in infrastructure maintenance and safety.

% The paper is structured as follows: Section \ref{sec:RWork} reviews the evolution of semantic segmentation techniques, focusing on the development of FPN and its applications. Section \ref{sec:method} details our methodology, including the design and implementation of our proposed E-FPN architecture. Section \ref{sec:experiments} presents the Culvert-Sewer Defects dataset and the benchmark Aerial Semantic Segmentation Drone dataset used to evaluate the model's performance. Section \ref{sec:implementation} outlines the implementation steps and metrics. Section \ref{sec:class_balance} discusses techniques for balancing the dataset. Section \ref{sec:results} provides an analysis of our experimental results and comparative evaluations against state-of-the-art methods. Section\ref{sec:DiscussionofFindings} discusses our findings, insights, and limitations. Finally, Section \ref{sec:conclusions} concludes the paper with a summary of our contributions and potential future research directions in infrastructure maintenance and safety.

\section{Related Work} \label{sec:RWork}

Semantic segmentation has significantly advanced with the development of deep learning techniques. This section provides a comprehensive review of these advancements, with a particular focus on FPNs and their applications in infrastructure inspection, especially for culverts and sewer pipes.

\subsection{Evolution of Semantic Segmentation Techniques}

Semantic segmentation has progressed from early methods based on hand-crafted features and conventional classifiers \cite{Zhou2021, Kushwah2021} to more advanced deep learning approaches. The advent of Convolutional Neural Networks (CNNs) marked a transformative shift, enabling more effective pixel-wise classification. A key milestone in this evolution was the introduction of Fully Convolutional Networks (FCNs) \cite{Benkert2021, Tian2022}, which facilitated dense predictions over arbitrary-sized inputs and laid the groundwork for subsequent semantic segmentation architectures \cite{Cen2021}. Among these advancements, encoder-decoder networks like U-Net \cite{ronneberger2015u}, bottom-up top-down networks such as FPNs \cite{lin2017feature}, and ViTs that leverage self-attention mechanisms for capturing long-range dependencies \cite{nguyen2024image} have become prominent in addressing various segmentation challenges.

In the following, we will highlight each of these network types in detail:

\subsubsection{Encoder-Decoder Architectures and U-Net Variants}

Encoder-decoder architectures, particularly U-Net \cite{ronneberger2015u}, have significantly advanced the field of semantic segmentation. U-Net's innovative use of skip connections enhances segmentation accuracy by effectively combining low-level and high-level features, which has proven especially beneficial in medical imaging.

Recent variants of U-Net have further refined its performance. For instance, Su et al. \cite{su2022research} integrated Convolutional Block Attention Modules (CBAM) into U-Net. This modification incorporates Channel Attention Modules (CAM) and Spatial Attention Modules (SAM), which collectively enhance the network's ability to focus on informative features and salient spatial regions, improving both global semantic understanding and local detail capture.

Further advancements include the Attention Sparse Convolutional U-Net (ASCU-Net) proposed by Tong et al. \cite{tong2021ascu}. ASCU-Net introduces a tripartite attention mechanism that combines Attention Gates (AG), Spatial Attention Modules (SAM), and Channel Attention Modules (CAM). This approach specifically targets important structures and emphasizes crucial spatial and channel-wise information, resulting in state-of-the-art segmentation accuracy across various domains.

\subsubsection{FPNs and Multi-Scale Feature Representation}
Bottom-up top-down networks, like FPN, have showcased their versatility in object detection and semantic segmentation tasks. FPN addresses the challenge of multi-scale feature extraction by constructing a hierarchical pyramid of feature maps with varying resolutions. By integrating contextual information at different scales, FPN facilitates more robust and accurate segmentation. Its utilization of both bottom-up and top-down pathways allows for effective feature fusion across different levels of abstraction, making it well-suited for tasks with significant variations in scale and orientation.

FPN typically utilizes pretrained networks in the bottom-up pathway. Tsung-Yi Lin et al. \cite{lin2017feature} applied FPN to object detection, using a ResNet backbone to extract hierarchical features from input images. The bottom-up pathway of the FPN starts with a standard ResNet architecture pretrained on the ImageNet dataset \cite{deng2009imagenet}. ResNet generates a hierarchy of feature maps at various scales. These feature maps are then processed through the top-down pathway, where higher-resolution features are obtained by upsampling spatially coarser but semantically stronger feature maps from the higher levels of the pyramid.

The integration of both bottom-up and top-down pathways allows for effective feature fusion across different levels of abstraction. This enables FPN to capture both global context and fine-grained details in the input data, making it well-suited for tasks with significant variations in scale and orientation, such as semantic segmentation of complex scenes or, in our case, culvert and sewer pipe defect detection. 

\subsubsection{ViT}
ViTs are a novel approach to image processing that diverge from traditional convolutional neural networks. Introduced by Dosovitskiy et al. \cite{nguyen2024image}, ViTs adapt the self-attention mechanisms from natural language processing to handle visual data. This allows them to capture long-range dependencies and global context in images effectively. While ViTs excel at understanding overall image features, they are computationally demanding and typically require extensive datasets for optimal performance. The Swin Transformer  \cite{liu2021Swin} enhances the ViT model by incorporating hierarchical features and local window-based attention. This design improves efficiency and scalability, overcoming limitations of standard ViTs. Using a shifted windowing scheme across multiple stages, Swin Transformer captures multi-scale features while maintaining computational efficiency. Its hierarchical structure effectively handles various object scales and feature resolutions, making it particularly suitable for complex tasks like semantic segmentation. The Swin Transformer’s ability to capture both local and global features, combined with its efficient handling of large-scale data, positions it as a promising architecture for infrastructure inspection tasks. Its performance in segmentation tasks, including those involving diverse and intricate features, suggests potential benefits for applications like culvert and sewer pipe defect detection.

\subsection{Research gap and motivation}

EDNs are effective at semantic segmentation but struggle with varying object scales. FPNs excel in handling multi-scale objects but may not address class imbalance as effectively as EDNs. FPNs also present high model complexity. ViTs offer promising performance but come with their own set of drawbacks. They generally require substantial computational resources and extensive training data, which can be challenging with smaller datasets. Additionally, ViTs may struggle with fine-grained details and local features due to their lack of inherent inductive biases, which can hinder performance in tasks that require high spatial resolution.

Our culvert-sewer defect dataset presents unique challenges due to its inherent characteristics, diversity, and class imbalance. Class imbalance adversely affects the performance of EDNs and FPNs. An EDN that excels in precise localization may struggle with diverse object scales, while an FPN might be biased towards overrepresented classes, despite its ability to handle scale variations. Similarly, while ViTs show promise, their high computational demands and potential difficulties with fine details make them less suitable for our specific needs.

The challenges in this domain, including small structural defects, varying pipe materials and sizes, and highly imbalanced defect classes, necessitate specialized approaches. Existing solutions have made progress in semantic segmentation and addressing data imbalance, but there is a gap in tailored solutions for culvert and sewer inspection.

Given these limitations, directly applying existing EDN, FPN, or ViT architectures is not ideal for our dataset. We need to explore alternative approaches that improve object segmentation and manage class imbalance without adding computational overhead. Our work introduces an E-FPN designed specifically for culvert and sewer defect segmentation. The E-FPN incorporates enhanced blocks with reduced complexity for efficient multi-scale feature extraction and integrates architectural improvements inspired by recent advancements in attention mechanisms. This approach addresses the multi-scale nature of defects and the challenge of class imbalance, aiming to develop a more robust and accurate segmentation model tailored to underground infrastructure inspection.

\section{Proposed method: E-FPN}\label{sec:method}
This section introduces the E-FPN, our proposed architecture for semantic segmentation in culvert and sewer pipe inspection. We describe the E-FPN's structure and key innovations, followed by a detailed ablation study. This study quantifies the impact of each architectural modification, supporting our design choices with empirical evidence. E-FPN builds on the foundational principles of traditional FPNs, incorporating innovative enhancements designed to address the specific difficulties encountered in this domain. The E-FPN is structured around two core components:

\begin{figure}[!ht]
    \centering
    \includegraphics[width=0.99\linewidth]{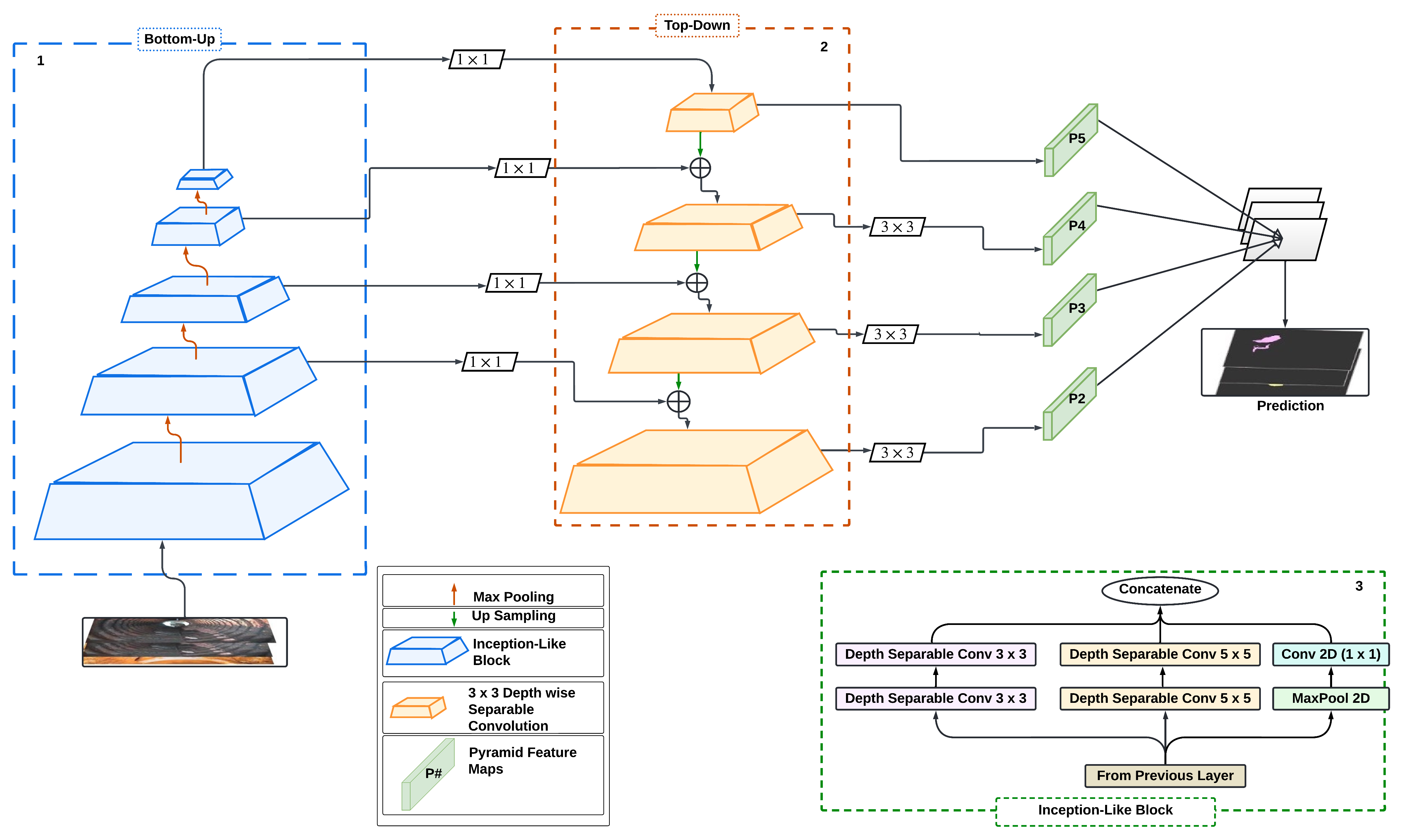}
 \caption{E-FPN architecture: Dual-pathway design for multi-scale feature extraction. The bottom-up pathway filters and down-samples the input image by a factor of 2 at each layer using enhanced Inception blocks with depth-wise separable convolutions. The top-down pathway employs upsampling and feature fusion to reconstruct a colored-masked image. Numbers indicate feature map dimensions and channel depths at each stage.}

    \label{fig:EFPN}
\end{figure}

\begin{enumerate}
    \item \textbf{Bottom-up pathway:} The bottom-up pathway forms the foundation of our E-FPN architecture, as illustrated in the blue dotted block in Figure \ref{fig:EFPN}. This pathway is responsible for extracting multi-scale features from the input image through a series of convolutional operations and downsampling stages.
    
    To enhance the effectiveness of this pathway for detecting defects in culverts and sewer systems, we implemented several key design modifications. Specifically, we replaced the standard bottom-up pathway layers with a custom Inception-like block, highlighted in the green dotted block in Figure \ref{fig:EFPN}. This custom block incorporates both \(3 \times 3\) and \(5 \times 5\) filters, along with a parallel max-pooling layer. The choice of these filter sizes was motivated by the characteristics of culvert and sewer imagery. The \(3 \times 3\) filters capture fine-grained details and textures crucial for identifying defects like hairline cracks or early-stage corrosion. These filters help detect small-scale anomalies indicating structural issues. The larger \(5 \times 5\) filters help capture larger-scale defects, such as joint misalignments and significant deformations, by providing a wider receptive field. This allows the network to understand the overall structure and condition of the pipe.
    
    Additionally, we incorporated two extra spatial detection layers compared to traditional Inception blocks. These layers enhance the model's capacity to learn and localize complex features specific to culvert and sewer defects, providing a more robust multi-scale feature representation. This is particularly valuable for detecting defects of varying sizes and appearances, depending on the camera's distance and angle.
    
    The additional layers improve gradient flow during backpropagation, leading to better learning of fine-grained features essential for identifying structural anomalies. This adaptation helps our custom Inception-like block effectively address the challenges of culvert and sewer defect detection, enabling more accurate analysis across different scales and defect types.
    
    Throughout the pathway, we used depth-wise separable convolutions to reduce the parameter count without compromising performance. The pathway begins with 64 filters and doubles after each max-pooling operation, balancing network capacity with efficiency to capture increasingly complex features while maintaining a manageable parameter count.

\item \textbf{Top-down pathway:} The top-down pathway, shown in the orange dotted block in Figure \ref{fig:EFPN}, enhances the bottom-up process by upsampling and merging features to create higher-resolution images. This maintains spatial details and allows for accurate defect localization. The pathway is designed using principles from deep learning, computer vision, and signal processing, specifically for inspecting culverts and sewer pipes.

Key aspects of the top-down pathway include:
\begin{itemize}
    \item Feature fusion: The pathway starts with a \(1 \times 1\) convolution on the final bottom-up layer to reduce channel depth to 128. Each previous layer is upsampled by a factor of two and merged with the corresponding bottom-up feature map. This fusion low-level and high-level features, crucial for accurate semantic segmentation. The \(1 \times 1\) convolution reduces dimensionality while retaining essential information, and the upsampling recovers spatial details lost during downsampling.

    \item \textbf{Aliasing mitigation:} To preserve fine details and sharp transitions, a \(3 \times 3\) depth-wise separable convolution is applied to all merged layers. This approach mitigates aliasing effects during upsampling and ensures high fidelity in the final segmentation output. The \(3 \times 3\) depth-wise separable convolution acts as a learnable anti-aliasing filter, which efficiently removes high-frequency artifacts.

     \item \textbf{Consistent output configuration:} A common classifier is shared across all output feature maps, maintaining a 128-dimensional output channel configuration. This uniform representation aids in the final segmentation task and ensures consistent defect recognition regardless of scale. This approach promotes scale-invariant feature learning, which is critical for accurate defect detection across varying camera distances.

     \item \textbf{Efficient upsampling:} Depth-wise separable convolutions in the upsampling process ensure computational efficiency while generating high-resolution feature maps. This design principle, demonstrated by Chollet \cite{chollet2017xception} in the Xception architecture, enables detailed segmentation without significant computational overhead.
     
 \end{itemize}

The design of the top-down pathway integrates feature fusion across abstraction levels, improving the network's ability to capture global context and fine-grained details. This is useful for culvert and sewer inspections as it facilitates accurate semantic segmentation. E-FPN offers a robust and efficient solution for semantic segmentation in these inspections by combining architectural innovations with data balancing strategies (explained in section \ref{sec:class_balance}), ensuring accurate segmentation of common and rare defect types.

\end{enumerate}

%%%%%%%%%%%%%%%%%%%%%%%%

\subsection{Progressive Enhancement of FPN Architectures: A Path to E-FPN}

This subsection provides a detailed examination of the modifications made to the original FPN to enhance various aspects of feature extraction and representation. We conducted extensive experiments on the model, leading to the evolution of our proposed E-FPN.

\begin{itemize}
\item \textbf{Original FPN with ResNet Backbone:} The baseline model uses the original FPN architecture with a ResNet backbone. This model serves as the foundation for evaluating subsequent modifications, establishing a reference point for performance comparisons.
\item \textbf{FPN with Atrous Convolutions:} We attempted to enhance the FPN architecture with atrous (dilated) convolutions to expand the receptive field while maintaining spatial resolution. Atrous convolutions insert gaps between kernel elements \cite{chen2017deeplab}, enabling the model to capture broader context and preserve fine details without increasing computational complexity. Atrous convolutions have improved semantic segmentation in models like DeepLab \cite{chen2017deeplab} by enhancing multi-scale contextual understanding. However, our experiments did not show significant performance gains from integrating atrous convolutions with FPN. This unexpected result may stem from compatibility issues between atrous convolutions and FPN, or from dataset-specific factors that didn't fully utilize the technique's advantages.
\item \textbf{FPN with Attention Gates:} Attention gates (AGs) were incorporated into the FPN to enhance feature prioritization. Introduced by Oktay et al. \cite{oktay2018attention}, AGs dynamically emphasize significant regions while suppressing less relevant ones. This mechanism improves the model's ability to differentiate between important and trivial features, leading to better segmentation performance. AGs work by learning weights that adaptively highlight relevant features in feature maps, guiding the network to focus on informative areas and ignore noise. In our experiments, integrating AGs into the network's paths significantly improved accuracy in identifying and segmenting critical features. The enhanced ability to highlight key regions and suppress less useful information resulted in more precise segmentation, particularly in complex and cluttered scenes. This approach proved effective in improving the FPN's performance for segmenting culvert and sewer pipe defects.

\item \textbf{FPN with Self-Attention Mechanisms:}  We explored integrating self-attention mechanisms into the FPN architecture, inspired by Transformer models' success \cite{rahman2024enhancing}. Self-attention allows the network to prioritize relevant input parts, capturing long-range dependencies and global context. Despite its potential to enhance feature relationships and contextual understanding, our experiments showed no significant performance improvement. This limited impact may be due to increased computational demands or dataset characteristics that don't fully leverage self-attention advantages. While promising in other contexts, self-attention proved less effective for our specific segmentation task.

\item \textbf{FPN with Squeeze-and-Excitation Blocks:} The enhanced Squeeze-and-Excitation (SE) block is an attention mechanism that improves channel-wise feature responses through adaptive recalibration. It compresses feature maps into a channel descriptor, summarizing global information for each channel. This descriptor then recalibrates feature responses, emphasizing important features and suppressing less relevant ones. By capturing channel interdependencies, SE blocks enhance the network's focus on crucial features, improving overall performance \cite{zhang2024reconstruction}.

We incorporated an enhanced version of SE blocks \cite{alshawi2023dual} into our feature extraction paths. This version includes a learnable recalibration rate, further refining the dynamic recalibration process. The integration aimed to improve information flow between bottom-up and top-down pathways. These SE blocks dynamically recalibrate channel-wise feature responses based on learned attention maps, significantly enhancing feature representation and segmentation accuracy. This improvement demonstrates the effectiveness of enhanced SE blocks in refining feature extraction and producing more precise segmentation results.

\item \textbf{FPN with Inception and Residual Blocks:} Integrating Inception blocks and residual connections into models architectures significantly enhances multi-scale feature extraction and supports the training of deeper networks \cite{rastogi2024multi, anand2024enhanced}. The Inception blocks capture features at multiple scales concurrently, while residual connections alleviate gradient issues, leading to substantial improvements in performance and robustness. In this experiment, we replaced FPN botton-up and top-down pathways blocks with Inception block with residual connections. The new model shows 10\% improvement over the original FPN. In this experiment, we replaced the bottom-up and top-down pathway blocks in FPN with Inception blocks featuring residual connections. The new model demonstrated a 15\% improvement over the original FPN.  

\item \textbf{FPN with Factorized Inception Block:} To reduce the complexity of the previous experiment, we used a factorized Inception block, which optimizes by decomposing large convolutions into smaller, more manageable operations \cite{szegedy2016rethinking}. This modification achieves a balanced trade-off between computational load and model performance, leading to notable improvements over the original FPN. 

\item \textbf{FPN with $5\times5$ Factorized Convolution:} The addition of a $5\times5$ factorized convolution was aimed at enhancing the model's feature extraction capabilities by expanding the receptive field. This adjustment successfully improved the model’s ability to capture and process features, leading to better overall performance. Both factorized versions of the Inception block improved network performance over the original FPN. However, using the unfactorized Inception block yielded higher performance compared to the factorized versions.

\item \textbf{FPN with Additional $1\times1$ Layer:} Introducing an additional 1x1 layer to the convolutional block was designed to further refine feature representation while maintaining computational efficiency. This enhancement contributed to improved feature extraction and model effectiveness, offering a more precise and nuanced analysis of the input data \cite{he2016deep}. However, our experiments show that adding the additional layer to the block results in a performance drop of at least 10\%.

\item \textbf{E-FPN (Proposed Model):} From these experiments, we observed that incorporating multi-scale block, like Inception block, significantly improves the performance of FPN but also increases computational overhead. Based on this observation, we developed our proposed E-FPN model, which integrates an advanced multi-scale block with reduced complexity, as detailed in the section above.
\end{itemize}

This study reveals that while several modifications to the FPN architecture contributed to performance improvements, the E-FPN model, with its combination of advanced features and optimizations, offers the most substantial gains in accuracy and robustness for semantic segmentation tasks, All these results are detailed in Section  \ref{sec:ablationResult}.  
%%%%%%%%%%%%%%%%%%%%%%%%%%%%%%%
\section{Datasets}\label{sec:experiments}

This section is  divided into two subsections. Section \ref{sec:Culvert-SewerDefects} describes the creation of the Culvert-Sewer Defects dataset, while Section \ref{sec:AerialDataset} discusses the benchmark Aerial Semantic Segmentation Drone Dataset used to assess model's accuracy.

\subsection{Culvert-Sewer Defects Dataset}\label{sec:Culvert-SewerDefects}

In this subsection, we detail the acquisition and preprocessing of source videos, which include various defect instances. We also outline the pixel-wise annotation strategy employed to create precise ground truth masks for semantic segmentation, culminating in our dataset of 6,300 images. The process is outlined as follows: 

\subsubsection{Data Collection and Class Importance Weights}\label{sec:CIW}

We collected 580 annotated videos of underground infrastructure inspections from two sources: the U.S. Army Corps of Engineers (USACE) and an industry partner. These videos cover culverts and sewer pipes and include a diverse range of real-world conditions, introducing variations in materials, shapes, dimensions, and imaging environments. This variety offers a comprehensive representation of typical inspection scenarios.

Skilled technicians reported the majority of the videos, identifying deficiencies in culverts or sewer pipes by type and location. This precise reporting facilitated the identification and annotation process for our task. A professional civil engineer assigned importance weights to each deficiency class, reflecting their economic and safety impacts based on U.S. industry standards. These weights were normalized to establish priorities during the learning process, as shown in Table \ref{tab:ninedefects}. These class importance weights (CIW) are used to measure the severity of each deficiency using the Frequency Weighted Intersection over Union (FWIoU) metric. The dataset encompasses a wide range of materials, shapes, and measurements found in culverts and sewer pipes, mirroring real-life inspections. This diversity presents a challenge due to the need to integrate data from various sources and structures.
\begin{comment}
\begin{table}[ht]
    \centering
    \caption{Culvert-sewer inspection classes: Deficiency and corresponding class importance weights (CIW).}
    \label{tab:ninedefects}
    \begin{tabular}{lc}
        \toprule
        \textbf{Deficiency} & \textbf{CIW} \\
        \midrule
        Water Level & 0.0310 \\
        Cracks & 1.0000 \\
        Roots & 1.0000 \\
        Holes & 1.0000 \\
        Joint Problems & 0.6419 \\
        Deformation & 0.1622 \\
        Fracture & 0.5100 \\
        Encrustation/Deposits & 0.3518 \\
        Loose Gasket & 0.5419 \\
        \bottomrule
    \end{tabular}
\end{table}
\end{comment}
%
\begin{table}[!ht]
\centering
\caption{Culvert-sewer inspection classes: Deficiency and corresponding class importance weights (CIW).}
\label{tab:ninedefects}
\resizebox{0.5\columnwidth}{!}{%
\centering
\begin{tabular}{|c||c|}
\hline
\textbf{Deficiency} & \textbf{CIW} \\

\hline
Cracks & 1.0000 \\
\hline
Roots & 1.0000 \\
\hline
Holes & 1.0000 \\
\hline
Joint Problems & 0.6419 \\
\hline
Deformation & 0.1622 \\
\hline
Fracture & 0.5100 \\
\hline
Water Level & 0.0310 \\
 \hline
Encrustation/Deposits & 0.3518 \\
\hline
Loose Gasket & 0.5419 \\

\hline
\end{tabular}
}
\end{table}

\subsubsection{Pixel-Wise Annotation for Semantic Segmentation Task}

We compiled our dataset by partitioning each video into frames captured at intervals of 4 to 10 seconds at key classification points within the culvert and sewer inspection footage. During the manual annotation process, we referred to inspection reports to identify the location and type of each deficiency, assigning pixel-wise annotations in specific colors according to the US NASSCO’s Pipeline Assessment Certification Program (PACP) guidelines \cite{pottawatomie}. Frames without any deficiencies were discarded.

\begin{figure}
    \centering
    \includegraphics[width=0.7\linewidth]{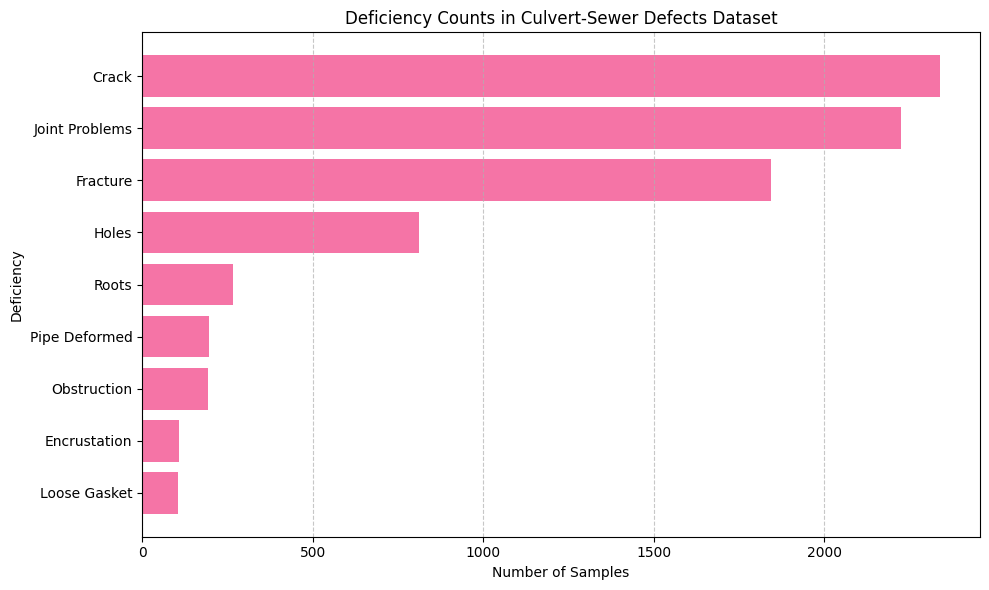}
\caption{Deficiency distribution in the Culvert-Sewer Defects Dataset. The dataset exhibits significant imbalance, with sample counts ranging from 2,340 in the highest class to 104 in the lowest class.}

    \label{fig:distribution}
\end{figure}

Each annotation is timestamped to the exact second of the video and linked to a specific deficiency class. Additionally, the pipe location for each annotation was recorded. The final dataset includes approximately 6,300 annotated frames, covering the nine deficiencies listed in Table \ref{tab:ninedefects} and illustrated in Figure \ref{fig:distribution}.

As shown in the figure, our dataset exhibits a significant class imbalance, with certain deficiencies, such as cracks, being more prevalent than others, such as roots. This imbalance poses challenges for modeling and is a key aspect of the dataset's characteristics.

\subsection{Aerial Semantic Segmentation Drone Dataset}\label{sec:AerialDataset}

The Aerial Semantic Segmentation Drone Dataset is designed to enhance the safety of autonomous drone flight and landing by providing detailed semantic annotations of urban scenes\cite{drone_dataset}. This dataset features high-resolution images captured from a bird's eye view at altitudes ranging from 5 to 30 meters, with each image measuring 6000x4000 pixels.
The dataset is divided into 400 images training set and 200 images for testing. 
It includes pixel-accurate annotations across 22 classes, such as tree, grass, dirt, water, person, car, and obstacle. Additionally, the dataset offers high-resolution RGB images, fish-eye stereo images, thermal images, and 3D ground truth data for specific scenes.
Although not officially labeled as a benchmark, the dataset's comprehensive annotations and diverse image types make it a valuable resource for evaluating semantic segmentation models. 

We employed this challenging dataset to demonstrate the robustness and efficiency of our model, confirming its ability to generalize across different tasks and validate its effectiveness in real-world applications.

\section{Imbalance Handling Techniques}\label{sec:class_balance}

The Culvert-Sewer Defects dataset exhibits significant class imbalance, with some defect types having a substantially larger number of samples compared to others ranging from 2,340 samples for some classes to as few as 104 samples for others, as shown in Figure \ref{fig:distribution}. This imbalance poses a challenge for model training and may lead to biased predictions, especially favoring the overrepresented classes. In semantic segmentation tasks, such imbalance can result in poor generalization and accuracy, particularly for minority classes \cite{li2020analyzing}. We explore two techniques to mitigate the effect of such imbalance on the model's performance: class decomposition and data augmentation.

\subsection{Class Decomposition and Ensemble Learning}\label{sec:classdecomp}

Class decomposition is used to address the imbalance issue by breaking down the multi-class segmentation problem into smaller tasks. This involves splitting the dataset into groups based on each class' characteristics and sample distribution \cite{johnson2019survey}. We organize the data into groups consisting of three classes each, based on the deficiency type features and available samples. For instance, if two classes share similar characteristics, like crack and fracture, we assign them to separate groups to prevent confusion during model training. This simplifies the task for the models and improves their pattern-learning ability.

We train our E-FPN model separately on smaller balanced datasets. After training, we combine the predictions using ensemble learning techniques.

Ensemble learning is a powerful approach where multiple models are combined to enhance overall performance. This method involves training several models, each of which may focus on different subsets of the data or specialize in particular classes. The strength of ensemble learning lies in its ability to aggregate the diverse insights and predictions from these individual models, leading to improved accuracy and robustness \cite{zhou2012ensemble}.

In our approach, ensemble learning plays a crucial role in leveraging the collective knowledge from multiple models. By aggregating their predictions, we can significantly enhance our model's robustness and generalization ability. This technique helps mitigate individual model weaknesses and provides a more reliable and stable performance, ultimately leading to more accurate and dependable results.
We evaluate the enhanced model using the entire test dataset. Our experiments show a significant improvement in model performance, validating the effectiveness of our approach in addressing dataset imbalance and enhancing predictive accuracy. Detailed results with these performance metrics are presented in Section \ref{sec:results}.

\subsection{Data Augmentation and Sampling Techniques}

Data augmentation is a pivotal technique in deep learning for improving model performance, particularly when dealing with imbalanced datasets. By introducing variations and diversifications into the training data, data augmentation helps models generalize better and become more robust to real-world scenarios \cite{khan2023review}. The key augmentation methods used and their impact on model training are outlined below:

\begin{itemize}
\item Horizontal Flip: This technique involves flipping images horizontally, effectively doubling the dataset size while maintaining the original class distribution. It helps the model generalize better by providing mirrored versions of images.
\item Gaussian Blur: Applying Gaussian blur introduces a smooth, blurred effect to images. This reduces overfitting to specific details and encourages the model to focus on more generalized features, enhancing its robustness.
\item Color Jittering: Color jittering involves randomly adjusting the brightness, contrast, saturation, and hue of images. This adds significant diversity to the dataset and helps the model learn features that are invariant to color variations, improving its adaptability.
\item Shearing: Shearing applies a shear transformation to images, distorting them along one axis. This introduces variability in object orientations, helping the model recognize objects from different perspectives.
\item Rotation: Rotating images by specified angles exposes the model to various object orientations, aiding in generalization to unseen views and enhancing overall robustness.
\item Random Noise: Adding random variations to pixel values simulates real-world image noise. This technique improves the model's robustness by exposing it to noisy data, helping it perform better in diverse conditions.
\item Random Crop: Randomly cropping portions of images forces the model to focus on different regions of interest, enhancing its ability to localize objects and improve detection accuracy.
\end{itemize}

These techniques are applied to each class in the training set to ensure a balanced representation. Additionally, for classes with more than 2000 samples, such as the joint problems class, under-sampling is used by randomly removing excess samples to achieve a more balanced class distribution. This combined approach of data augmentation and strategic sampling is crucial for improving model performance and generalization ability.
\begin{figure}[!ht]
    \centering
    \includegraphics[width=0.95\linewidth]{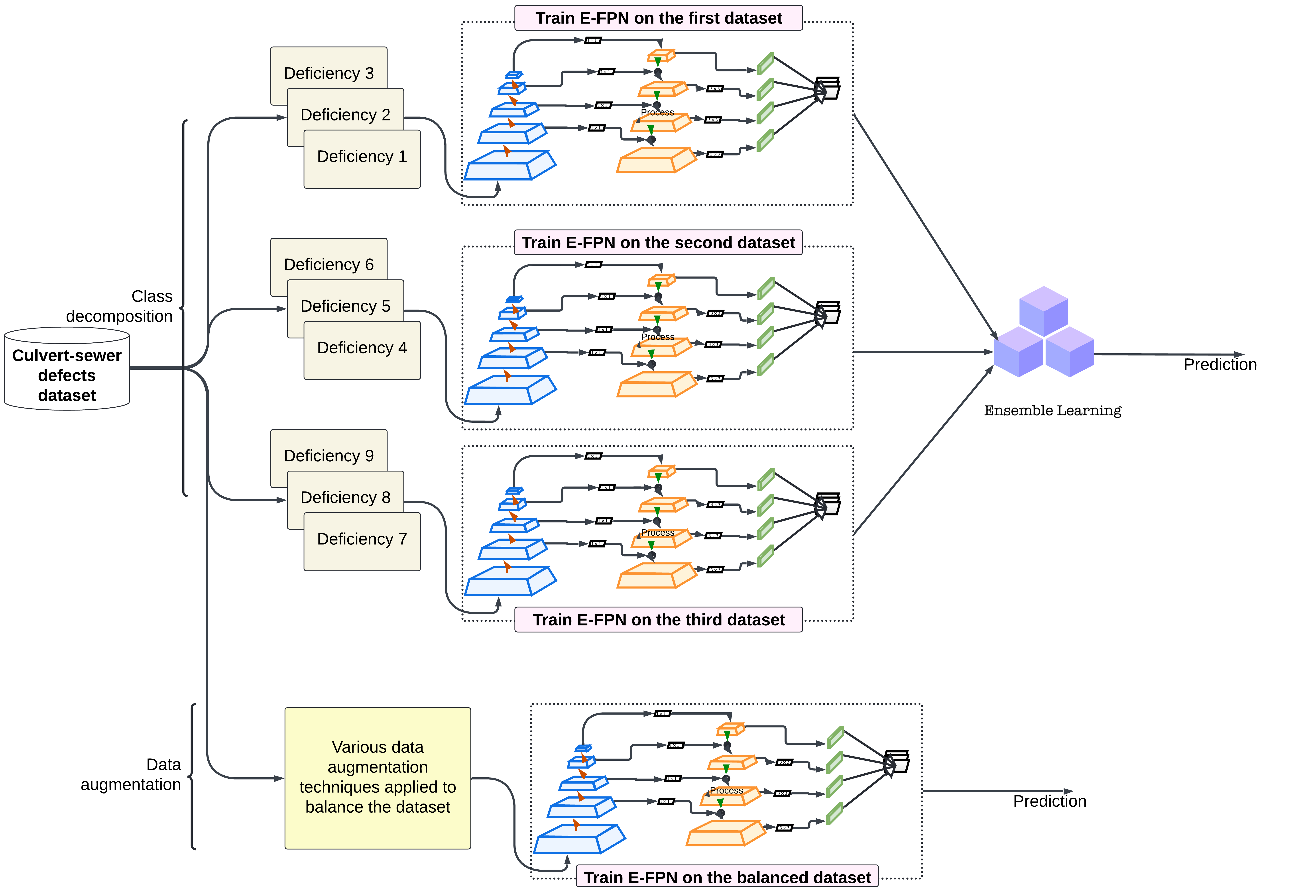}
\caption{Workflow for Mitigating Class Imbalance through Class Decomposition and Data Augmentation Techniques. The figure illustrates the process of applying class decomposition to group similar classes and the  targeted data augmentation to balance the dataset. Models trained on these balanced samples have improved performance and generalization.}

    \label{fig:aug-dec}
\end{figure}

\subsection{Combining Class Decomposition and Data Augmentation}

To enhance model performance, we integrate class decomposition and data augmentation techniques. Class decomposition simplifies the multi-class segmentation problem by grouping classes with similar sample sizes into clusters. Despite this, minor imbalances may still exist within these clusters. To further enhance balance within each cluster, we apply targeted data augmentation techniques. After training the models on these more balanced clusters, we combine predictions using ensemble learning, as detailed in Section \ref{sec:classdecomp}. This combined strategy ensures that each cluster benefits from diverse and balanced training examples, ultimately leading to improved model performance and generalization.

Our analysis of the combined approach shows a remarkable improvement in performance, demonstrating its effectiveness. Figure \ref{fig:aug-dec} illustrates the workflow of class decomposition and data augmentation, showing the steps from dataset preparation to model evaluation.

%%%%%%%%%%%%
\section{Experimental Setup}\label{sec:implementation}

In this section, we describe the methodologies and parameters used for developing, training, and evaluating our semantic segmentation model. We cover the optimization strategies, loss functions, evaluation metrics, and other key aspects of the implementation.

\paragraph{Optimization and Loss Functions}
For training our semantic segmentation models, we employ the Adam optimizer with an initial learning rate of 0.001. Adam is chosen for its efficiency in handling sparse gradients and dynamically adjusting learning rates. The loss function used is Categorical Cross-Entropy, which is effective for multi-class pixel classification tasks by minimizing the difference between predicted probabilities and ground-truth labels.

\paragraph{Evaluation Metrics}
We evaluate model performance using the following metrics: 1. Intersection over Union (IoU): Measures segmentation accuracy by comparing predicted and ground-truth masks. 2. Frequency-Weighted IoU (FWIoU): Accounts for class frequencies using CIW as explained in \ref{sec:CIW}. 3. F1 Score: Balances precision and recall, useful for imbalanced datasets. 4. Balanced Accuracy: Averages recall across classes, effective for imbalanced datasets. 5. Matthews Correlation Coefficient (MCC): Assesses classification quality in skewed datasets.
% We utilize a diverse set of metrics to evaluate model performance:
% \begin{itemize}
% \item \textbf{Intersection over Union (IoU):} Measures the spatial overlap between predicted and ground-truth segmentation masks, providing insight into segmentation accuracy.
% \item \textbf{Frequency-Weighted IoU (FWIoU):} Takes into account class frequencies, offering accurate understanding of class representation. For class frequencies, we used CIW explained in section \ref{sec:CIW}
% \item \textbf{F1 Score:} The harmonic mean of precision and recall, providing a balanced measure of a model's accuracy by considering both false positives and false negatives, particularly useful for imbalanced datasets.
% \item \textbf{Balanced Accuracy:} Averages recall across all classes, useful for evaluating performance in imbalanced datasets.
% \item \textbf{Matthews Correlation Coefficient (MCC):} Assesses classification quality, particularly in skewed datasets.
% \end{itemize}

\paragraph{Training Procedures}
Models are trained for 100 epochs on the Culvert-Sewer Defects dataset and the Aerial Semantic Segmentation Drone Dataset. The datasets are split into training (70\%), validation (15\%), and test (15\%) subsets to evaluate generalization performance. Baseline models are also established and evaluated under the same conditions for comparison.

\paragraph{Hardware and Software}
Training is conducted on NVIDIA T4 GPUs using Keras with TensorFlow, providing the computational power and tools necessary for efficient model training and evaluation.

\section{Results}\label{sec:results}
We evaluated the effectiveness of E-FPN against state-of-the-art semantic segmentation architectures. Additionally, we tested E-FPN on the Aerial Semantic Segmentation Drone dataset to demonstrate its robustness and adaptability to diverse imagery types. This section is organized into three subsections: Subsection \ref{sec:baseLineModels} provides a comprehensive comparison with state-of-the-art models, including quantitative metrics and visualizations. It highlights E-FPN's effectiveness in addressing multi-scale feature representation across both the Culvert-Sewer Defects and Aerial Semantic Segmentation Drone datasets. Subsection \ref{sec:Impact of Data Imbalance Mitigation Techniques} discusses the impact of data imbalance mitigation techniques on model performance, detailing the effects of class decomposition and data augmentation. Subsection \ref{sec:ablationResult} presents a detailed ablation study, analyzing the contribution of individual components and architectural modifications to E-FPN's performance.

\subsection{Comparison with Baseline Architectures}\label{sec:baseLineModels} 

To evaluate the efficiency of our proposed E-FPN, we compared it with several state-of-the-art semantic segmentation architectures, including the original FPN, U-Net, CBAM-enhanced U-Net, ASCU-Net, and Swin Transformer, as shown in Table \ref{tab:Performance2}.

The original FPN model used in our comparison was built on a ResNet backbone, pretrained on the ImageNet dataset to leverage learned features, and then fine-tuned on our specific dataset. This fine-tuning process tailored the model to the characteristics and requirements of our dataset.

The Swin Transformer used in this experiment is integrated with the UPerNet framework for semantic segmentation. Specifically, the "upernet-Swin-small" model, which is hosted on Hugging Face, combines the Swin Transformer with UPerNet's components, including a Feature Pyramid Network (FPN) and a Pyramid Pooling Module (PPM). This integration enhances the model's ability to capture multi-scale features and context for improved segmentation performance. The Swin Transformer was pretrained on the ImageNet dataset, providing it with general feature representations before being fine-tuned for the semantic segmentation task.

We evaluated Swin Transformer models under two conditions: with and without pretraining. In the non-pretrained condition, the models were trained from scratch on our dataset. This allowed us to evaluate the impact of pretraining on the performance of both Swin models and the original FPN.

\begin{comment}
\begin{table*}[ht]
\centering
\caption{Performance comparison of various models on culvert-sewer defects dataset (w/bg: with background, w/o bg: without background)}
\label{tab:Performance2}
\resizebox{\textwidth}{!}{%
\begin{tabular}{
  @{}
  l
  S[table-format=1.5]
  S[table-format=1.5]
  S[table-format=1.5]
  S[table-format=1.5]
  S[table-format=1.5]
  S[table-format=1.5]
  @{}
}
\toprule
\textbf{Model} & {\textbf{IOU w/ bg}} & {\textbf{IOU w/o bg}} & {\textbf{FWIoU}} & {\textbf{F1}} & {\textbf{Bal. Acc}} & {\textbf{MCC}} \\
\midrule
FPN with ResNet (original) & 0.69947 & 0.66575 & 0.69657 & 0.80610 & 0.81387 & 0.83922 \\
U-Net & 0.58559 & 0.53906 & 0.48980 & 0.69333 & 0.63078 & 0.40457 \\
CBAM U-Net & 0.60501 & 0.55889 & 0.67053 & 0.71269 & 0.67964 & 0.71296 \\
ASCU-Net & 0.70358 & 0.67021 & 0.71491 & 0.81161 & 0.79463 & 0.79451 \\
Swin Transformer & 0.58130 & 0.53244& 0.62829 & 0.66690 & 0.70790 & 0.69966 \\
E-FPN (this paper) & 0.77187 & 0.74601 & 0.78073 & 0.86346 & 0.84264 & 0.85005 \\
\bottomrule
\end{tabular}
}
\end{table*}

\end{comment}
%

\begin{table}[!ht]
\centering
\caption{Performance comparison of various models on culvert-sewer defects dataset (w/bg: with background, w/o bg: without background)}
\label{tab:Performance2}
\resizebox{1.0\columnwidth}{!}{%
\centering
\begin{tabular}{|c||c||c||c||c||c||c|}
\hline
\textbf{Model} & {\textbf{IOU w/ bg}} & {\textbf{IOU w/o bg}} & {\textbf{FWIoU}} & {\textbf{F1}} & {\textbf{Bal. Acc}} & {\textbf{MCC}} \\
\hline
 FPN with ResNet (original) & 0.69947 & 0.66575 & 0.69657 & 0.80610 & 0.81387 & 0.83922 \\
 \hline
U-Net & 0.58559 & 0.53906 & 0.48980 & 0.69333 & 0.63078 & 0.40457 \\
\hline
CBAM U-Net & 0.60501 & 0.55889 & 0.67053 & 0.71269 & 0.67964 & 0.71296 \\
\hline
ASCU-Net & 0.70358 & 0.67021 & 0.71491 & 0.81161 & 0.79463 & 0.79451 \\
\hline
Swin Transformer & 0.58130 & 0.53244& 0.62829 & 0.66690 & 0.70790 & 0.69966 \\
\hline
\textbf{E-FPN (this paper)} & \textbf{0.77187} & \textbf{0.74601 }& \textbf{0.78073} & \textbf{0.86346} & \textbf{0.84264} & \textbf{0.85005} \\

\hline
\end{tabular}
}
\end{table}

\begin{figure}[!ht]
    \centering
    \includegraphics[width=0.99\linewidth]{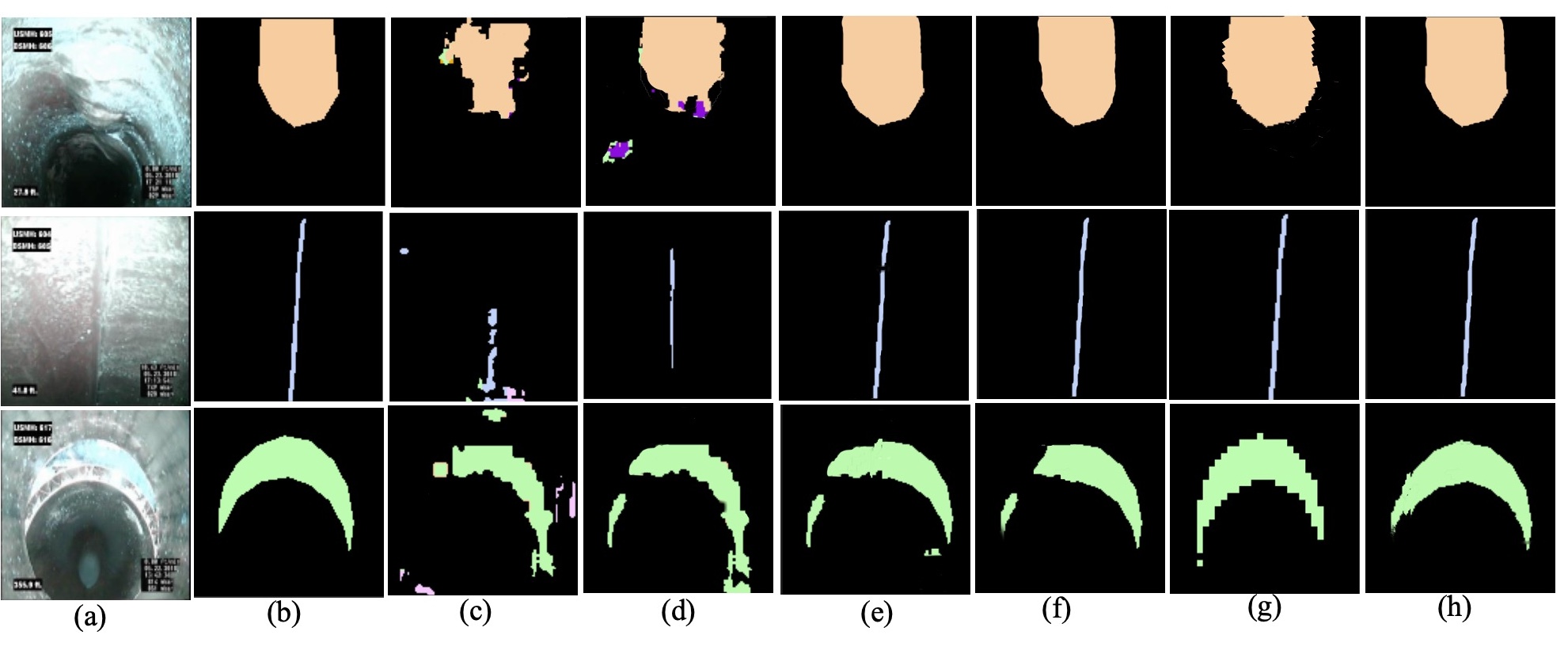}
    \caption{ Comparative segmentation results on the culvert-sewer defects dataset are shown, with the first row illustrating pipe deformation defects, the second row showing a crack, and the third row illustrating joint misalignment: (a) Original images (b) Ground truth (c) U-Net (d) CBAM U-Net (e) FPN with ResNet (f) ASCU-Net (g) Swin Transformer, (h) E-FPN (this paper)}
    \label{fig:modelsVR}
\end{figure}

\begin{figure} [!ht]
    \centering
    \includegraphics[width=0.99\linewidth]{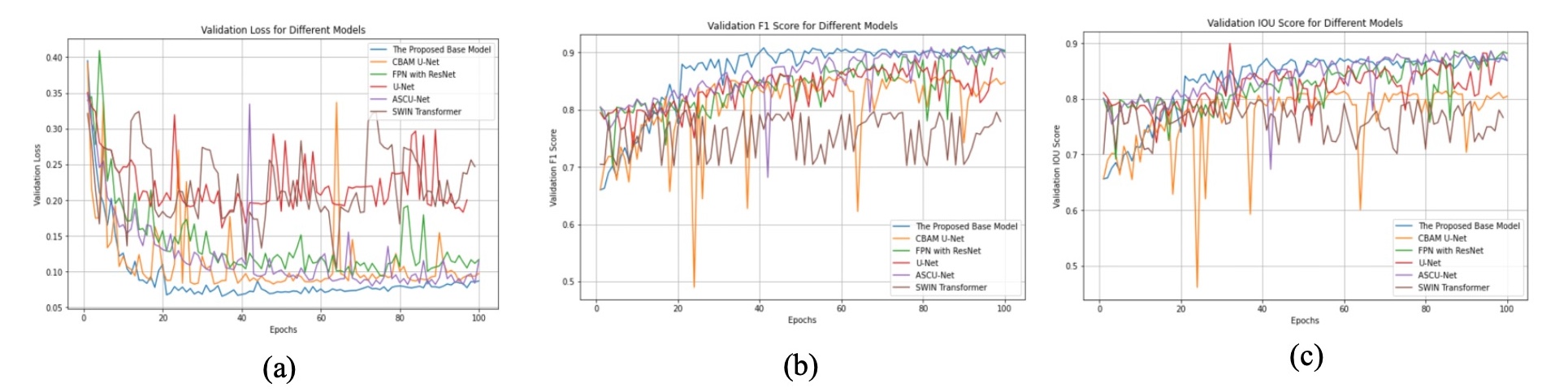}
    \caption{Comparative validation metrics of E-FPN against baseline and state-of-the-art models on the culvert-sewer defects dataset: (a) Cross-entropy loss, (b) F1-score, and (c) IoU. The proposed model in the blue color shows the highest validation IoU and F1-score compared to the other models. }

    \label{fig:graph-label}
\end{figure}

Figure \ref{fig:modelsVR} presents a visual comparison of the models evaluated in our study, highlighting their reconstruction capabilities. U-Net and CBAM U-Net show limitations in accurately reconstructing images. Although these models successfully identify deficiencies, they struggle to fully represent the fine details, leading to incomplete reconstructions. In contrast, the Swin model, which uses shifted window self-attention mechanisms, exhibits visual artifacts in its outputs. These artifacts could be due to the challenges in capturing complex details with the Swin model's hierarchical attention approach. Our model, however, demonstrates superior performance by capturing and representing fine details more effectively, resulting in more accurate and visually coherent reconstructions. This is quantitatively demonstrated by an average IoU improvement of 13.8\% over other models on the culvert-sewer defects dataset. Figure \ref{fig:graph-label} displays the validation graphs for these models on Culvert-Sewer Defects dataset.

In addition to our primary evaluations, we tested our proposed model on the Aerial Semantic Segmentation Drone dataset. For this comparison, we focused on three base models: the original U-Net, the original FPN, and our proposed model. This allowed us to benchmark our model against established baseline architectures. As shown in Table \ref{tab:PerformancAerial}, our proposed model consistently outperforms both the original U-Net and FPN across various metrics. Our model achieves an average IoU improvement of 27.3\% over these baseline models, demonstrating its effectiveness and strong performance on different datasets.

\begin{table}[!ht]
\centering
\caption{Performance Comparison of Various Models on Aerial Semantic Segmentation Drone Dataset. w/bg: with background, w/o bg: without background.}
\label{tab:PerformancAerial}

\resizebox{1.0\columnwidth}{!}{%

\begin{tabular}{|c||c||c||c||c||c|}
\hline
\textbf{Model} & \textbf{IOU w/ bg} & \textbf{IOU w/o bg}  & \textbf{F1} & \textbf{Bal. Acc} & \textbf{MCC} \\

\hline
 U-Net & 0.56269 &0.55617 & 0.63382 & 0.72782 & 0.62041 \\
\hline
FPN with ResNet (original)& 0.58375  & 0.57148 & 0.68216 &	0.73202 & 0.65047\\
\hline
\textbf{E-FPN (this paper)}  &\textbf{0.72937 }&	\textbf{0.71645} & \textbf{ 0.75059 }& \textbf{0.86504} & \textbf{0.76868} \\
\hline
\end{tabular}
}
\end{table}

\begin{figure} [ht]
    \centering
    \includegraphics[width=0.99\linewidth]{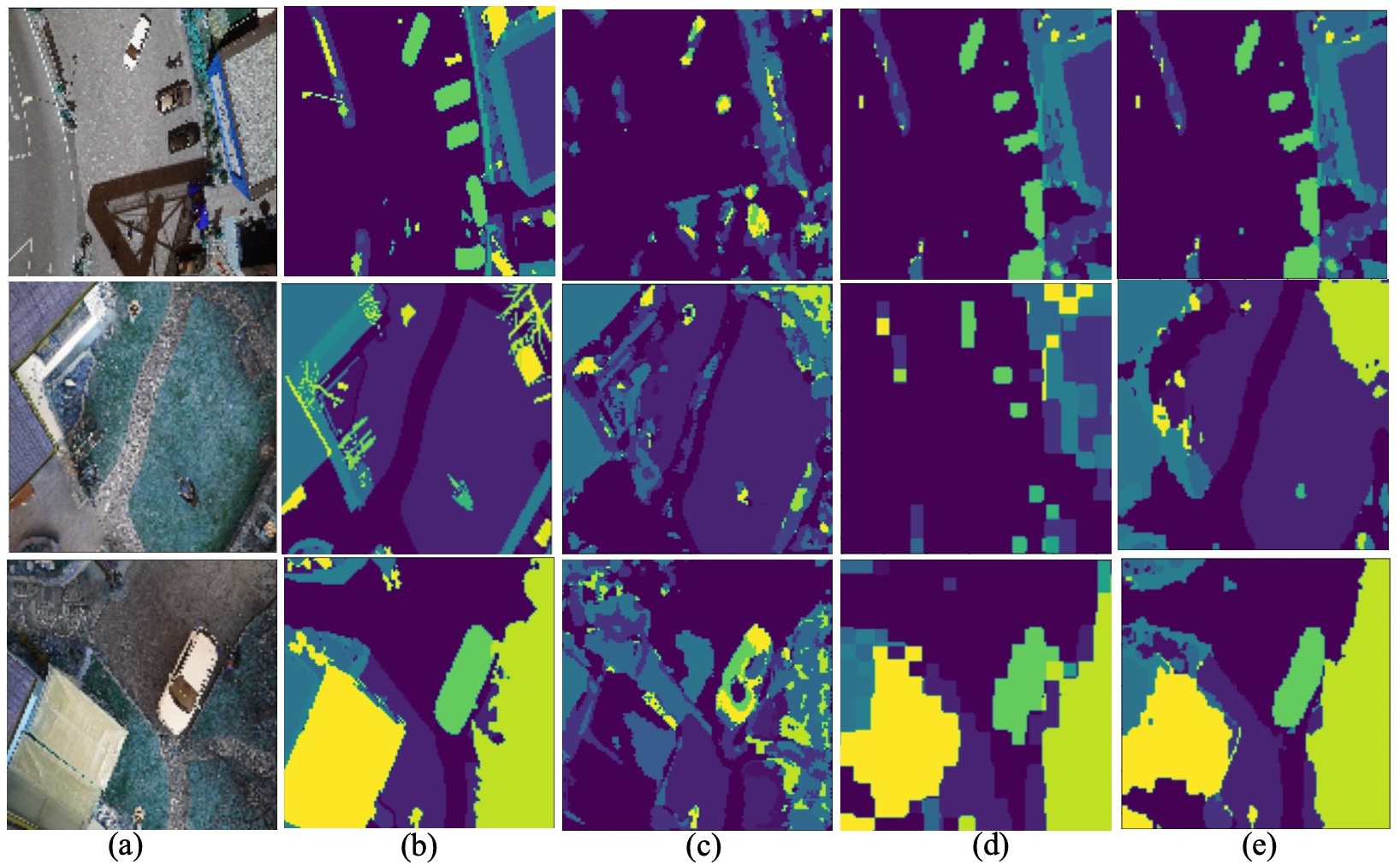}
  \caption{Comparative segmentation results on the Aerial Semantic Segmentation Drone Dataset are illustrated, featuring three samples with different types of trees, buildings, cars, and other classes. The results are presented as follows: (a) Original images, (b) Ground truth, (c) FPN with ResNet, (d) U-Net, and (e) The proposed E-FPN model.}\label{fig:Areial}

\end{figure}

Figure \ref{fig:Areial} illustrates that while there is a room for improvement across all models, our proposed model stands out by consistently exceeding the performance of baseline models. This is evident in various test cases from the dataset, where our model excels in accurately classifying and segmenting diverse classes, including trees, grass, dirt, water, people, cars, and obstacles. This comparison emphasizes the superior robustness and effectiveness of our model in managing intricate segmentation challenges, outperforming both the original U-Net and FPN models.

A standout feature of the comparative analysis is our model's exceptional efficiency, achieved with a significantly smaller parameter count compared to other architectures. Our model utilizes only 1.32 million parameters, resulting in reductions of approximately 19.40 times, 23.43 times, 23.57 times, 24.01 times, and 61.6 times compared to the original FPN, U-Net, CBAM U-Net, ASCU-Net, and Swin Transformer, respectively, as shown in Table \ref{tab:numberofparams}.

\begin{table}[!ht]
\centering
  \caption{Comparison of the number of trainable parameters in different models}
  \label{tab:numberofparams}%
   
\resizebox{0.9\columnwidth}{!}{%
\centering
\begin{tabular}{|c||c|}
\hline
  Model & Number of Trainable Parameters \\
\hline
    FPN \cite{lin2017feature} & 25,698,557 \\
    \hline
    U-Net \cite{ronneberger2015u} & 31,032,521 \\
    \hline
    CBAM U-Net \cite{su2022research} & 31,221,065 \\
    \hline
    ASCU-Net \cite{tong2021ascu} & 31,841,202 \\
    \hline
    Swin Transformer \cite{liu2021Swin} &  81,367,128\\
    \hline
   \textbf{ E-FPN (this paper)} & \textbf{1,324,660 }\\
\hline
\end{tabular}
}
\end{table}

%
\begin{comment}
\begin{table}[htbp]
  \centering
  \caption{Comparison of the number of trainable parameters in different models}
    \begin{tabular}{lc}
    \toprule
    Model & Number of Trainable Parameters \\
    \midrule
    FPN \cite{lin2017feature} & 25,698,557 \\
    U-Net \cite{ronneberger2015u} & 31,032,521 \\
    CBAM U-Net \cite{su2022research} & 31,221,065 \\
    ASCU-Net \cite{tong2021ascu} & 31,841,202 \\
    Swin Transformer \cite{liu2021Swin} &  81,367,128\\
    E-FPN (this paper) & 1,324,660 \\
    \bottomrule
    \end{tabular}%
  \label{tab:numberofparams}%
\end{table}%
\end{comment}

\subsection{Impact of Data Imbalance Mitigation Techniques on Model Performance}\label{sec:Impact of Data Imbalance Mitigation Techniques}
In this subsection, we discuss the results of applying two data imbalance mitigation techniques: class decomposition and data augmentation. These techniques were employed to enhance the model's performance on imbalanced datasets, with a specific focus on improvements observed for individual classes.
We begin by presenting the overall performance of the proposed E-FPN model under different data balancing techniques, as summarized in Table \ref{tab:model_performance}. This table compares the model's performance across various metrics, including IoU with background, IoU without background, FWIoU, F1 score, and balanced accuracy. These results provide a broad overview of how each technique impacts the model's general performance.

\begin{comment}
\begin{table*}[!ht]
    \centering
        \caption{Performance comparison of E-FPN for different data balancing techniques}
    \label{tab:model_performance}
    \resizebox{\textwidth}{!}{%
    \begin{tabular}{
        @{}
        l
        S[table-format=1.5]
        S[table-format=1.5]
        S[table-format=1.5]
        S[table-format=1.5]
        S[table-format=1.5]
        @{}
    }
        \toprule
        \textbf{Model} & {\textbf{IOU w/ bg}} & {\textbf{IOU w/o bg}} & {\textbf{FWIoU}} & {\textbf{F1}} & {\textbf{Bal. Acc}} \\
         & \multicolumn{1}{c}{\textbf{}} & \multicolumn{1}{c}{\textbf{}} & \multicolumn{1}{c}{\textbf{}} & \multicolumn{1}{c}{\textbf{}} & \multicolumn{1}{c}{\textbf{}} \\
        \midrule
        E-FPN trained on the full dataset & 0.77187 & 0.74601 & 0.78073 & 0.86346 & 0.84264 \\
        E-FPN trained using class decomposition & 0.81236 & 0.79084 & 0.76147 & 0.89959 & 0.85980 \\
        E-FPN trained using data augmentation & 0.80120 & 0.77938 & 0.86780 & 0.86271 & 0.87325 \\
        E-FPN trained using data augmentation and class decomposition & 0.82573 & 0.80731 & 0.78472 & 0.91548 & 0.89608 \\
        \bottomrule
    \end{tabular}%
    }
\end{table*}
\end{comment}

\begin{table}[!ht]
\centering
        \caption{Performance Comparison of E-FPN for Different Imbalance Mitigation Techniques}
    \label{tab:model_performance}
\resizebox{1.0\columnwidth}{!}{%
\centering
\begin{tabular}{|c||c||c||c||c||c||c|}
\hline
\textbf{Model} & {\textbf{IOU w/ bg}} & {\textbf{IOU w/o bg}} & {\textbf{FWIoU}} & {\textbf{F1}} & {\textbf{Bal. Acc}} & {\textbf{MCC}} \\
\hline
  E-FPN trained on the full dataset & 0.77187 & 0.74601 & 0.78073 & 0.86346 & 0.84264 & 0.85005  \\
  \hline
        E-FPN trained using class decomposition & 0.81236 & 0.79084 & 0.76147 & 0.89959 & 0.85980& 0.86158 \\
        \hline
        E-FPN trained using data augmentation & 0.80120 & 0.77938 & 0.86780 & 0.86271 & 0.87325& 0.87380 \\
        \hline
        \textbf{E-FPN trained using data augmentation and class decomposition} &\textbf{ 0.82573 }& \textbf{0.80731} & \textbf{0.78472} & \textbf{0.91548} & \textbf{0.89608} & \textbf{0.89222 }\\
\hline
\end{tabular}
}
\end{table}

Following the overall comparison, we explore the class-wise IoU scores to examine the specific impact on individual classes, as shown in Table \ref{tab:class_iou_comparison}. This table highlights the effectiveness of each technique in improving IoU scores for different defect classes, with a detailed analysis of the benefits of using class decomposition, data augmentation, and their combination.

\begin{table}[!ht]
    \centering
    \caption{Class-wise IoU comparison of E-FPN for different data balancing techniques (CD: class decomposition, DA: data augmentation).}
    \label{tab:class_iou_comparison}
    \resizebox{1.0\columnwidth}{!}{%
    \centering
    \begin{tabular}{|l||c||c||c||c|}
    \hline

        \multicolumn{1}{|c||}{\textbf{Class}} & \textbf{E-FPN IoU (Full Dataset)} & \textbf{E-FPN IoU (CD)} & \textbf{E-FPN IoU (DA)} & \textbf{E-FPN IoU (DA \& CD)} \\

    \hline
    Class 0 - Background & 0.97882 & 0.98454 & 0.97575 & 0.99157 \\
    \hline
    Class 1 - Crack & 0.53762 & 0.62115 & 0.47168 & 0.62857 \\
    \hline
    Class 2 - Hole & 0.95873 & 0.96988 & 0.96490 & 0.96985 \\
    \hline
    Class 3 - Root & 0.87741 & 0.85661 & 0.88980 & 0.90543 \\
    \hline
    Class 4 - Deformation & 0.79313 & 0.77772 & 0.77328 & 0.82814 \\
    \hline
    Class 5 - Fracture & 0.56659 & 0.65706 & 0.61292 & 0.63682 \\
    \hline
    Class 6 - Encrustation & 0.72817 & 0.86972 & 0.89169 & 0.90083 \\
    \hline
    Class 7 - Joint Problems & 0.76837 & 0.77796 & 0.73421 & 0.78870 \\
    \hline
    Class 8 - Loose Gasket & 0.73803 & 0.79659 & 0.89660 & 0.80766 \\
    \hline
    Class 9 - Obstruction & 0.76318 & 0.77401 & 0.76392 & 0.79980 \\
    \hline
    Average IoU w/ bg & 0.77187 & 0.81236 & 0.80120 & \textbf{0.82574} \\
    \hline
    Average IoU w/o bg & 0.74601 & 0.79084 & 0.77939 & \textbf{0.80731} \\
    \hline
    \end{tabular}
    }
\end{table}

The results show that applying both class decomposition and data augmentation leads to the most significant improvements, particularly in challenging classes. For instance, class decomposition alone resulted in a 5.24\% improvement in average IoU, while data augmentation yielded a 3.66\% increase. When combined, these techniques produced a 6.97\% enhancement, demonstrating their complementary effects.

\begin{figure}[!ht]
    \centering
    \includegraphics[width=0.99\linewidth]{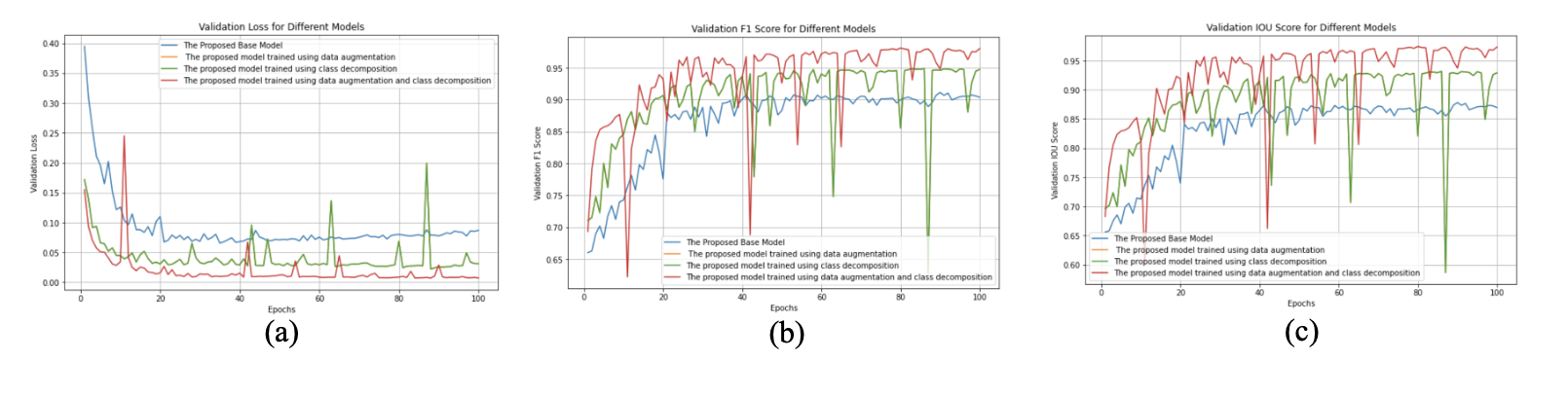}
    \caption{Comparative validation metrics for E-FPN under various data balancing techniques: (a) Cross-entropy loss, (b) F1-score, and (c) Intersection over Union (IoU). Results shown for the culvert-sewer defects dataset, with curves representing mean values over 5-fold cross-validation.}
    \label{fig:vgraph}
\end{figure}

To further illustrate these findings, Figure \ref{fig:vgraph} presents the validation graphs of these techniques, showing comparative metrics for cross-entropy loss, F1-score, and IoU. These visualizations reinforce the quantitative results by highlighting the consistent performance gains achieved through data balancing strategies.

E-FPN's enhanced performance, coupled with class decomposition and data augmentation, shows great promise for advancing infrastructure inspection techniques. The model's 6.97\% IoU enhancement using both techniques demonstrates their synergistic effect and the importance of addressing data imbalance in real-world applications, crucial in infrastructure maintenance to prevent catastrophic failures. The model's generalization across diverse scenarios suggests its potential applicability in various infrastructure inspection contexts, shown by its performance on the Aerial Semantic Segmentation Drone dataset. However, the computational overhead introduced by class decomposition and varying effectiveness of data augmentation across defect types present challenges for deployment in resource-constrained environments, highlighting the need for efficient, adaptive techniques to optimize performance while minimizing computational costs. 

\subsection{Ablation Study: Impact of Architectural Components on E-FPN Performance} \label{sec:ablationResult}

We conducted comparative analyses to evaluate the evolution of our E-FPN.  we focused on comparing different architectures within the FPN family. We compared various modifications of FPN, including the proposed E-FPN, against the baseline FPN model with a ResNet backbone. These variations encompassed FPN with atrous convolutions to enlarge receptive fields, FPN with AGs for enhanced feature selection, FPN incorporating Inception and residual blocks, self-attention mechanisms to capture long-range dependencies, enhanced SE blocks for channel-wise attention, and versions with factorized convolutions for computational efficiency. Each variant was designed to address specific aspects of feature extraction and representation. This analysis provides insights into how architectural choices impact performance in culvert and sewer defect segmentation. Table \ref{tab:Performance1} summarizes the results of the comparisons, presenting metrics including IoU, FWIoU, F1-Score, Balanced Accuracy, and MCC, for a detailed assessment of each architecture's strengths and limitations.

%%%

\begin{comment}
\begin{table*}[htbp]
\centering
\caption{Performance comparison of various FPN versions on Culvert-Sewer Defects dataset (w/bg: with background, w/o bg: without background)}
\label{tab:Performance1}
\resizebox{\textwidth}{!}{%
\begin{tabular}{
  l
  S[table-format=1.5]
  S[table-format=1.5]
  S[table-format=1.5]
  S[table-format=1.5]
  S[table-format=1.5]
  S[table-format=1.5]
}
\toprule
\textbf{Model} & {\textbf{IOU w/ bg}} & {\textbf{IOU w/o bg}} & {\textbf{FWIoU}} & {\textbf{F1}} & {\textbf{Bal. Acc}} & {\textbf{MCC}} \\
\midrule
FPN with ResNet (original) & 0.69947 & 0.66575 & 0.69657 & 0.80610 & 0.81387 & 0.83922 \\
FPN with Atrous Convolutions & 0.67904 & 0.64233 & 0.71527 & 0.79169 & 0.80171 & 0.81820 \\
FPN with Attention Gates & 0.75914 & 0.73176 & 0.77205 & 0.85471 & 0.81567 & 0.82716 \\
FPN with Self Attention & 0.64433 & 0.60399 & 0.67169 & 0.76452 & 0.71301 & 0.74522 \\
FPN with Enhanced Squeeze and Excitation Block & 0.74834 & 0.71971 & 0.76321 & 0.84697 & 0.82925 & 0.83254 \\
FPN with Inception and Residual Block & 0.74932 & 0.72081 & 0.76870 & 0.84705 & 0.82028 & 0.83331 \\
FPN with Factorized Inception Block & 0.71863 & 0.68662 & 0.74015 & 0.82435 & 0.78443 & 0.80874 \\
FPN with 5x5 Factorized Block & 0.72878 & 0.69784 & 0.75432 & 0.80056 & 0.79095 & 0.81078 \\
Adding 1x1 Layer to the Block & 0.68950 & 0.65404 & 0.73218 & 0.80359 & 0.79132 & 0.81690 \\
E-FPN (this paper) & 0.77187 & 0.74601 & 0.78073 & 0.86346 & 0.84264 & 0.85005 \\
\bottomrule
\end{tabular}
}
\end{table*}
\end{comment}

%
\begin{table}[!ht]
\centering
\caption{Performance comparison of various FPN versions on Culvert-Sewer Defects dataset (w/bg: with background, w/o bg: without background)}
\label{tab:Performance1}
\resizebox{1.0\columnwidth}{!}{%
\centering
\begin{tabular}{|c||c||c||c||c||c||c|}
\hline
\textbf{Model} & {\textbf{IOU w/ bg}} & {\textbf{IOU w/o bg}} & {\textbf{FWIoU}} & {\textbf{F1}} & {\textbf{Bal. Acc}} & {\textbf{MCC}} \\
\hline
 FPN with ResNet (original) & 0.69947 & 0.66575 & 0.69657 & 0.80610 & 0.81387 & 0.83922 \\
 \hline
FPN with Atrous Convolutions & 0.67904 & 0.64233 & 0.71527 & 0.79169 & 0.80171 & 0.81820 \\
\hline
FPN with Attention Gates & 0.75914 & 0.73176 & 0.77205 & 0.85471 & 0.81567 & 0.82716 \\
\hline
FPN with Self Attention & 0.64433 & 0.60399 & 0.67169 & 0.76452 & 0.71301 & 0.74522 \\
\hline
FPN with Enhanced Squeeze and Excitation Block & 0.74834 & 0.71971 & 0.76321 & 0.84697 & 0.82925 & 0.83254 \\
\hline
FPN with Inception and Residual Block & 0.74932 & 0.72081 & 0.76870 & 0.84705 & 0.82028 & 0.83331 \\
\hline
FPN with Factorized Inception Block & 0.71863 & 0.68662 & 0.74015 & 0.82435 & 0.78443 & 0.80874 \\
\hline
FPN with 5x5 Factorized Block & 0.72878 & 0.69784 & 0.75432 & 0.80056 & 0.79095 & 0.81078 \\
\hline
Adding 1x1 Layer to the Block & 0.68950 & 0.65404 & 0.73218 & 0.80359 & 0.79132 & 0.81690 \\
\hline
\textbf{E-FPN (this paper)} & \textbf{0.77187} & \textbf{0.74601 }& \textbf{0.78073} & \textbf{0.86346} & \textbf{0.84264} & \textbf{0.85005} \\

\hline
\end{tabular}
}
\end{table}

While some techniques, such as atrous convolutions and self-attention mechanisms, caused a drop in model performance, others led to significant improvements. Enhancements like the enhanced SE blocks and Additive AGs resulted in noticeable performance gains compared to the original FPN. Additionally, replacing the model's layers with Inception blocks that include residual connections yielded similar improvements.

However, factorized versions of the Inception block did not provide the expected benefits. Adding extra layers to these blocks also led to a performance drop. This outcome is consistent with the effect of self-attention mechanisms, highlighting that increasing model complexity can sometimes negatively impact performance, particularly on our specific Culvert-Sewer Defects dataset.

Consequently, we developed the E-FPN model, which incorporates a multi-scale depth-wise separable block. This block combines the advantages of varying filter sizes, similar to the Inception block, while achieving a reduction in computational complexity by at least 7 times fewer FLOPs. The proposed E-FPN model demonstrates a 10.35\% improvement over the original FPN and approximately 3.1\% improvement over the FPN models with Inception and attention mechanisms.

\section{Conclusions} \label{sec:conclusions}

E-FPN is an innovative semantic segmentation architecture that enhances the traditional FPN framework. It incorporates sparsely connected blocks and depth-wise separable convolutions to address data imbalance issues. The dual-pathway design and efficient convolution operations improve performance on imbalanced datasets while enhancing computational efficiency. E-FPN outperformed both traditional and state-of-the-art segmentation models when tested on our Culvert-Sewer Defect dataset and a benchmark aerial drone dataset. It achieved average IoU improvements of 13.8\% and 27.2\% respectively, surpassing FPN, U-Net, CBAM U-Net, ASCU-Net, and Swin Transformer.

E-FPN demonstrated superior performance on imbalanced datasets with improved generalization. Class decomposition and data augmentation increased IoU by 5.24\% and 3.66\% respectively, with a combined improvement of 6.9\%. E-FPN also achieved a 96.04\% reduction in model parameters compared to other evaluated models, showcasing its efficiency and versatility for real-world applications without compromising performance. 

Future research will focus on enhancing E-FPN's capabilities and applicability in real-world scenarios. Priorities include integrating temporal information from video streams for real-time detection, exploring unsupervised pre-training on large-scale unlabeled data to enhance feature extraction and generalization, and investigating active learning strategies and physics-informed neural networks for rare defect detection and domain knowledge incorporation. Efforts will be directed towards developing computationally efficient, adaptive techniques for resource-constrained environments, and expanding the model's applicability to diverse infrastructure inspection contexts. These advancements aim to establish E-FPN as a robust solution for automated infrastructure inspection, addressing challenges in data scarcity, computational efficiency, and critical defect detection.

\section*{Acknowledgments}
This research was partly supported by the U.S. Department of the Army, U.S. Army Corps of Engineers (USACE) under contract W912HZ-23-2-0004. The views expressed in this article are solely those of the authors and do not necessarily reflect the views of USACE.

\bibliographystyle{IEEEtran}
\bibliography{ref_EFPN}

\newpage
\vfill

\end{document}